\documentclass[11pt,a4paper]{article}
\usepackage[hyperref]{acl2021}
\usepackage{graphicx} 
\usepackage{xcolor}  %
\usepackage{times}
\usepackage{tabularx}
\usepackage{latexsym}
\usepackage{arydshln}
\usepackage{booktabs}
\usepackage{comment}
\usepackage{bbm}

\renewcommand{\UrlFont}{\ttfamily\small}

\usepackage{microtype}

\usepackage{amsmath,amsfonts,bm}

\def\eqref#1{equation~\ref{#1}}

\def\1{\bm{1}}

\def\mS{{\bm{S}}}

\DeclareMathAlphabet{\mathsfit}{\encodingdefault}{\sfdefault}{m}{sl}
\SetMathAlphabet{\mathsfit}{bold}{\encodingdefault}{\sfdefault}{bx}{n}

\def\sB{{\mathbb{B}}}
\def\sC{{\mathbb{C}}}
\def\sD{{\mathbb{D}}}

\def\sN{{\mathbb{N}}}

\def\sR{{\mathbb{R}}}

\aclfinalcopy %

\newcommand{\orange}[1]{{\color[HTML]{ff7f0e}\textbf{#1}}}

\newcommand{\green}[1]{{\color[HTML]{2ca02c}\textbf{#1}}}

\title{Biomedical Entity Linking with Contrastive Context Matching}

\author{
  Shogo Ujiie$^{\spadesuit}$ \quad Hayate Iso$^{\heartsuit}$\thanks{~~Work done while at Nara Institute of Science and Technology.} \quad Eiji Aramaki$^{\spadesuit}$ \\
$^\spadesuit$Nara Institute of Science and Technology  \quad
$^\heartsuit$Megagon Labs\\
\texttt{\{ujiie, aramaki\}@is.naist.jp}\\
\texttt{hayate@magagon.ai}}
\date{}

\begin{document}
\maketitle
\begin{abstract}
We introduce \textsc{BioCoM}, a contrastive learning framework for biomedical entity linking that uses only two resources: a small-sized dictionary and a large number of raw biomedical articles.
Specifically, we build the training instances from raw PubMed articles by dictionary matching and use them to train a context-aware entity linking model with contrastive learning.
We predict the normalized biomedical entity at inference time through a nearest-neighbor search. %
Results found that \textsc{BioCoM} substantially outperforms state-of-the-art models, especially in low-resource settings, by 
effectively using the context of the entities.
\end{abstract}

\section{Introduction}
Biomedical entity normalization, often referred to as entity linking in general domains, is the task of mapping entity mentions to unified concepts in a biomedical knowledge graph.
This is useful preprocessing task for many downstreaming tasks (e.g., information extraction~\cite{Lee2016-fu} and relation extraction~\cite{Xu2016-ha}).
This task is challenging as similar-looking words can have different meanings and same concepts can have different surface forms.
For example, while ``TDP-43 proteinopathy" and ``TdP" appear to be similar, they refer to different concepts.
``TDP-43 proteinopathy" is a generic term for diseases caused by abnormality of ``TDP-43" protein in the nervous system (e.g., amyotrophic lateral sclerosis),  but ``TdP" is a type of arrhythmia.

The approach used to normalize biomedical entities involves the linking entity mentions to the most similar synonym (and the corresponding concept) defined in the dictionary, where similarity is determined using representations of entities and synonyms.
Various approaches, such as supervised learning on a large amount of labeled dataset~\cite{Mondal2019-yk,Sung2020-ra,noauthor_undated-kc, Ujiie2021-gq} and self-supervised learning on a large dictionary~\cite{Liu2020-mr}, have been introduced to obtain
representations of entities.
Although these methods have achieved high performance, they use large-scale, difficult-to-obtain resources (e.g., manually annotated datasets or a large-sized dictionary).

\begin{figure}[t]
  \begin{center}
    \includegraphics[clip,width=7.0cm]{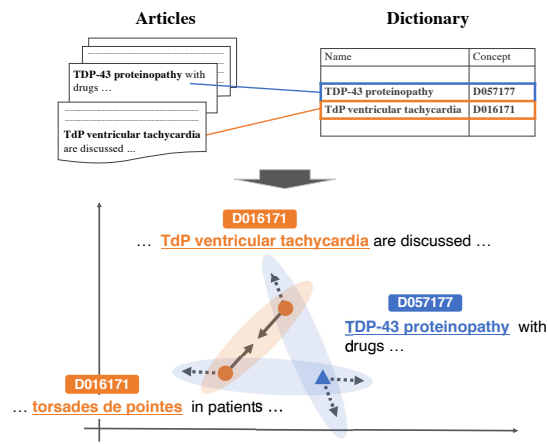}
    \caption{Overview of our framework, \textsc{BioCoM}. We build the training instances from raw biomedical articles by dictionary matching and use it to train the context-aware biomedical entity normalization model with contrastive learning.}
    \label{fig:overview}
  \end{center}
\end{figure}

This study aims to develop a biomedical entity normalization methodology for diseases that is effective even for limited resource scenarios.
Specifically, our work is under the following assumptions: unavailability of manually annotated data and provision of access only to a small-sized dictionary (i.e., dictionary where the number of synonyms for each concept is small).
These are practical assumptions in medical domain studies as biomedical concept annotations are costly because domain knowledge is required.
Note that diseases are some of the most important biomedical entities.
Therefore, this study, similar to most studies conducted by the biomedical text processing community, focuses on the normalization of disease mentions.~\cite{Leaman2013-lm,Lou2017-fd}.

This study proposes \textsc{BioCoM}, a contrastive learning framework for context-aware biomedical entity normalization.
\textsc{BioCoM} only uses a small-sized dictionary and a large number of raw biomedical articles.
Figure \ref{fig:overview} illustrates the overview of \textsc{BioCoM}.
As training instances to consider the context of the entities, we construct the corpus of the entity-linked sentences, i.e., a set of sentences that contain synonyms in the dictionary, from raw biomedical articles.
Accordingly, we show that the context obtained from the corpus strongly benefits the entity-linking model.
\textsc{BioCoM} can obtain cues for linking from the words surrounding the entities, such as synonyms that are not in the dictionary but are present in the sentence.
Thus, it works well even when the size of the dictionary is small.

The following are the contributions of this study: 
({\romannumeral 1}) introduction of a framework for biomedical entity normalization
that only requires a small-sized dictionary and a large amount of raw biomedical articles
({\romannumeral 2}) execution of the evaluation for limited-resource settings without manually annotated dataset or other large external resources
({\romannumeral 3}) achievement of state-of-the-art performance on three biomedical entity linking datasets, particularly significant in limited-resource scenarios.
The code is available at {\UrlFont https://github.com/ujiuji1259/biocom}.

\section{Related Work}
Distributed representations of entities play an important role in biomedical entity normalization tasks ~\cite{Mondal2019-yk,Li2017-nn,noauthor_undated-kc}.
Recently, bidirectional encoder representations from transformer (BERT) models~\cite{Devlin2019-kj} pre-trained on large biomedical documents have shown their effectiveness in biomedical entity normalization tasks.
For example, BioSyn~\cite{Sung2020-ra}, which uses BioBERT~\cite{Lee2020-ne} to obtain entity representations and fine-tunes them using synonym marginalization, has outperformed many existing methods in biomedical entity normalization tasks.
However, this approach requires a large annotated dataset for training.

Conversely, \textsc{SapBert}~\cite{Liu2020-mr} uses only a large biomedical knowledge graph, rather than an annotated dataset, for fine-tuning the entity representation and yields state-of-the-art performance in the biomedical entity normalization task.
However, \textsc{SapBert}'s performance is highly dependent on the size of the knowledge graph, and we have experimentally confirmed that the normalization performance degrades significantly when available resources are limited.

Our framework, \textsc{BioCoM}, could be assumed to be in line with \textsc{SapBert}, which does not use an annotated training dataset.
However, \textsc{BioCoM} further considers the context of entities using contextual entity representations derived from entity-linked sentences.

\section{Methods}
We propose a contrastive learning framework called \textsc{BioCoM} for context-aware biomedical entity normalization without any manually annotated datasets or large-scaled dictionaries.
``Context-aware" indicates that \textsc{BioCoM} considers both  entity mentions and entire sentences as input.
Formally, given a sentence $X$ and $i$-th entity mention $x_i$, we predict the corresponding concept $y_i \in \displaystyle \sC$, where $\displaystyle \sC$ is a set of concepts in the dictionary.
In this framework, input entity mentions and synonyms are encoded into the contextual representations.
These representations are learned
from entity-linked sentences' corpus 
through contrastive learning~\cite{he2020momentum,pmlr-v119-chen20j}.
At inference time, the input entity is normalized by context matching
from all the entities in the corpus.
We start with the corpus construction process and then the training procedure and inferences.

\subsection{Entity-Linked Sentences}
To obtain the synonyms' context in the dictionary, the corpus, a set of entity-linked sentences wherein the entities (synonyms in this case) are linked to the corresponding concepts, is required.
Here, 
the corpus
is built from raw biomedical articles by dictionary matching against synonyms (Figure \ref{fig:overview}).
We assume that we can access a database $\displaystyle \sD = \{(x_j^{'(m)}, y_j^{'(m)})\}, m=1\ldots M, j=1\ldots |x^{'(m)}|$, where $(x_j^{'(m)}, y_j^{'(m)})$ is the $j$-th entity mention and concept pair in a sequence of mention and concept pairs $(x^{'(m)}, y^{'(m)})$ in the $m$-th sentence.
Please refer to Section \ref{detail}, discussing the details of the corpus used in our experiments.

\subsection{Contrastive Learning}
\label{Training}
To conduct context matching, the entities in the database should be embedded into the feature space where entities with the same concept are close to each other.
Hence, we adopt contrastive learning, which is a framework used to learn similar/dissimilar representations from the similar/dissimilar pairs.
In this study, the positive (similar) and negative (dissimilar) pairs come from the mini-batch of the 
sentences.
That is, given an entity in the mini-batch, entities with the same concept within the mini-batch are treated as positive samples, while others are negative samples.

Regarding the loss function,
we use Multi-Similarity loss~\cite{Wang2019-id}, which is a metric-learning loss function that considers relative similarities between positive and negative pairs.
Let us denote the set of entities in the mini-batch by $\displaystyle \sB$ and the set of positive and negative samples for the entity $x'_i \in \displaystyle \sB$ by $\mathcal{P}_i$ and $\mathcal{N}_i$.
We define the cosine similarity of two entities $x'_i$ and $x'_j$ as $S_{i,j}$, resulting in a similarity matrix $\displaystyle \mS \in \displaystyle \sR^{|\displaystyle \sB| \times |\displaystyle \sB|}$.
Based on $\mathcal{P}_i$, $\mathcal{N}_i$, and $\displaystyle \mS$, the following training objectives are set:
\begin{equation*}
\begin{split}
	\label{eq-MS}
	\mathcal{L}_{MS} = \frac{1}{|\displaystyle \sB|}\sum_{i=1}^{|\displaystyle \sB|} \bigg\{\frac{1}{\alpha}  { \log \big[1 + \sum_{k  \in \mathcal{P}_i } e^{-\alpha (S_{ik} - \lambda)}}\big]  \\
	+ \frac{1}{\beta }  { \log \big[1+ \sum_{k \in \mathcal{N}_i}
		 e^{\beta (S_{ik} - \lambda)} \big]} \bigg\},
\end{split}   
\end{equation*}
where $\alpha, \beta$ are the temperature scales and $\lambda$ is the offset applied on $\displaystyle \mS$.
For pair mining, we follow the original paper~\cite{Wang2019-id}.

\subsection{Context Matching}
\label{model}
Our approach is to identify a concept $y$ for the entity mention $x$ by context matching from the %
database.
Specifically, the problem can be defined as the $K$-nearest neighbor search~\cite{fix1951discriminatory} using the contextual entity representations.
The neighbors $\displaystyle \sN_K (x, \displaystyle \sD)$ are obtained from
the database 
based on the cosine similarity with the contextual representations of the target entity mention $x$.
During inference, we predict $\hat{y}$ to be the concept that is the most frequent in $\displaystyle \sN_K (x, \displaystyle \sD)$.

\section{Experiments}
\subsection{Experimental Setting}
\label{detail}
\paragraph{Resource}
Herein, we use MEDIC~\cite{Davis2012-pp} as the dictionary .
MEDIC lists 13,063 diseases  from OMIM and MeSH and have over 70,000 synonyms linked to the concepts.
To limit the dictionary's size, we randomly sampled half of the synonyms for each concept.
Finally, the dictionary used in our experiments has three synonyms (on average) for each concept.

PubMed abstracts were used to construct entity-linked sentences.
The articles that appear in the test dataset were filtered out, obtaining in approximately 100M sentences.
The synonyms in MEDIC that appear in the sentences are then linked to the corresponding concepts.
We chose the longest match for any two overlapping synonyms appeared in sentences.

\paragraph{Test Datasets \& Evaluation Metric}
We evaluated \textsc{BioCoM} on three datasets for the biomedical entity normalization task: NCBI disease corpus (NCBID)~\cite{Dogan2014-nv}, BioCreative V Chemical Disease Relation (BC5CDR)~\cite{Li2016-gz}, and MedMentions~\cite{Mohan2019-tx}.
Following previous studies~\cite{DSouza2015-hi,Mondal2019-yk}, we used the accuracy as the evaluation metric.

Given that BC5CDR and MedMentions contain mentions whose concepts are not in MEDIC, these were filtered out during the evaluation.
We refer to these as ``BC5CDR-d" and ``MedMentions-d" respectively.

\paragraph{Model Details}
The contextual representation for each entity $x$ was obtained from PubMedBERT~\cite{pubmedbert}, which was trained on a large number of PubMed abstracts using BERT~\cite{Devlin2019-kj}.
Specifically,
the entity tokens were wrapped with special tokens, [\textit{ENT}] and [\textit{/ENT}], to mark the beginning and end of the mention, and feed this modified sequences to PubMedBERT.
We use the representations of [\textit{ENT}] as contextual representations~\cite{soares2019matching} of the entity that follows. Diagram for this entity representations can be found in Supplementary Materials.

The number of the nearest neighbors was set to $k = 15$ as it yielded the highest performance in the development set of MedMentions-d.

\subsection{Results}
\begin{table}
\centering
\small
    \begin{tabular}{lccc}
    \toprule 
    Methods & NCBID & BC5CDR-d & MedMentions-d\\
    \midrule
    TF-IDF & 45.3 & 59.9 & 57.8\\
    PubMedBERT & 34.6 & 42.2 & 45.2 \\
    \textsc{SapBert} & 56.5 & 65.2 & 65.1 \\
    \textsc{BioCoM} & \textbf{60.3} & \textbf{69.0} & \textbf{71.4} \\
    \bottomrule
    \end{tabular}
\caption{\label{result} Accuracy on three datasets.
Note that the dictionary and target concepts that are used for training and inference are different from previous work~\cite{Liu2020-mr}.}
\end{table}

\begin{table}[t!]
    \centering
    \footnotesize
    \begin{tabularx}{\linewidth}{lX}
        \toprule 
        Label: \green{D014178} & Input: ... phenotype B-I for \textit{B-ALL}, phenotype ... the presence of \textbf{translocation t(4;11))}\\
        \hdashline
        \textsc{SapBert} & [\orange{D054868}] \textbf{partial 11q monosomy syndrome} \\
        \textsc{BioCoM} & [\green{D014178}] ... involved in \textbf{genetic translocations} characteristic of b-cell acute lymphoblastic leukemia (\textit{B-cell ALL}). \\
        
        \midrule
        Label: \green{D002292} & Input: High occurrence of \textbf{non-clear cell renal cell carcinoma} \\
        \hdashline
        \textsc{SapBert} & [\green{D002292}] \textbf{renal cell carcinomas}\\
        \textsc{BioCoM} & [\orange{D002289}] ... used to treat \textit{renal cell} \textbf{carcinoma, non-small-cell lung} cancer and colon cancer ... \\
        \bottomrule
\end{tabularx}
\caption{\label{example} Predicted samples from \textsc{SapBert} and \textsc{BioCoM} in MedMentions-d.
The [predicted concept] and the nearest neighbor are shown. %
Entity mentions and the cue phrases are written in \textbf{bold} and \textit{italic}, respectively.}
\end{table}

We compared the performance of \textsc{BioCoM} with three baseline models: tf-idf, PubMedBERT~\cite{pubmedbert}, and \textsc{SapBert}~\cite{Liu2020-mr}.
All these models did not require manually annotated datasets and only used entity mentions as inputs.
Please refer to the Supplementary Materials for the details of these models.

Table~\ref{result} shows that 
\textsc{BioCoM} obtains significant improvements over the baseline models for all the datasets.
This demonstrates that the context-matching approach -- based on the contextual entity representations learned from the automatically constructed corpus -- is effective for biomedical entity normalization.

\section{Discussion}
\subsection{Effect of the Dictionary Size}
To determine the effect of the number of synonyms, we performed experiments with different numbers of synonyms in the dictionary: 39,959, 86,205, 151,559, and 218,325.
To augment synonyms, we use ones from UMLS~\cite{Bodenreider2004-bh}, a large biomedical knowledge graph, and the remaining from MEDIC.

\begin{figure}
    \centering
    \includegraphics[clip,width=5.7cm]{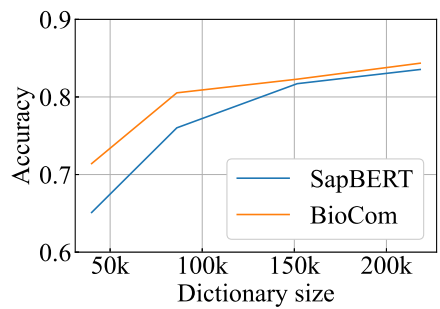}
    \caption{Effect of the number of synonyms on MedMentions-d.}
    \label{fig:dictsize}
\end{figure}

Figure \ref{fig:dictsize} illustrates the accuracy on MedMentions-d for each dictionary size.
This indicates that \textsc{BioCoM} yields even better results when the number of synonyms in the available dictionary is small while keeping superior results for the fully-expanded dictionary.

\subsection{Qualitative Analysis}
Here, we qualitatively analyze \textsc{BioCoM} to understand its behavior.
Table \ref{example} shows the predicted concepts and the nearest neighbors for input entity mentions in the MedMentions-d.

In the first example, \textsc{BioCoM} predicts the concept correctly while \textsc{SapBert} fails to find the correct answer.
This is because \textsc{BioCoM} utilizes the context as the cue to normalize the entity mention (``B-ALL" in the input sentence and ``B-cell ALL" in the nearest neighbor sentence).
This result strongly demonstrates the effectiveness of the context-matching approach and \textsc{BioCoM} can successfully learn the contextual entity representations by focusing on the cues within the sentences.

The second example shows an erroneous prediction from \textsc{BioCoM} owing to the annotation mistake.
We can determine that the entity mention ``carcinoma, non-small-cell lung" is wrongly recognized in place of ``renal cell carcinomas" in the nearest neighbor sentence.
This is because \textsc{BioCoM} uses the longest match algorithm to recognize entity mentions.
Thus, the fact that "carcinoma, non-small-cell lung" is longer yields a match. 
This causes a serious mismatch, wherein \textsc{BioCoM} successfully focuses on ``renal cell carcinomas" to encode the entities, but the linked mention is "carcinoma, non-small-cell lung."
This mismatch can be addressed by adopting named entity recognition system if available.

\section{Conclusion}
We introduced \textsc{BioCoM}, a contrastive learning framework for context-aware biomedical entity normalization without manually annotated datasets or large-sized dictionaries.
Our experiments on three datasets showed that our model significantly outperformed state-of-the-art models, especially when the available resources are limited.

\bibliography{acl2020}
\bibliographystyle{acl_natbib}

\appendix
\section{Data Sampling Strategy}
For every mini-batch, we randomly chose a certain number of concepts and randomly sampled two sentences, including a synonym of the concept.
This was performed so that at least one positive pair exists for each concept within the mini-batch.
Here, we chose 16 concepts for each mini-batch, thus obtaining 32 unique sentences per mini-batch.

\section{Dataset Details}
The dataset used in our experiments is threefold: the NCBI disease corpus (NCBID)~\cite{Dogan2014-nv}, the BioCreative V Chemical Disease Relation (BC5CDR) task corpus ~\cite{Li2016-gz}, and MedMentions~\cite{Mohan2019-tx}.
We list the number of documents and disease mentions in each dataset in Table \ref{dataset}.

\begin{table}
\centering
\small
\begin{tabular}{lcc}
\toprule 
dataset & \# of documents & \# of mentions \\
\midrule
NCBID & 100 & 960\\
BC5CDR & 500 & 4,424 \\
MedMentions & 879 & 3,795 \\
\bottomrule
\end{tabular}
\caption{\label{dataset} Number of documents and entity mentions in each dataset.}
\end{table}

\paragraph{NCBID}
NCBI disease corpus consists of 793 PubMed titles and abstracts, each of which contains manually annotated disease mentions.
They are manually separated into training (593), development (100), and test (100) sets.
Each disease mentions are mapped into the CUI that is contained in the MEDIC dictionary ~\cite{Davis2012-pp}.

\paragraph{BC5CDR}
BC5CDR is the task of chemical-induced relation extraction comprising 1500 PubMed abstracts.
Abstracts are equally separated into training, development, and test sets.
BC5CDR provides manually annotated disease and chemical entities, which are mapped into MEDIC and comparative toxicogenomics database (CTD) chemical dictionaries, respectively.
We only used disease entities linked to the concepts in MEDIC.

\paragraph{MedMentions}
MedMentions is the set of large-scale annotated resources used to recognize the biomedical concepts.
It provides more than 4,000 PubMed abstracts and over 350,000 mentions linked to the concepts in unified medical language system (UMLS).
For our experiments, we converted concept unique identifiers in UMLS to the entities in MEDIC using UMLS.
As UMLS contains many varieties of medical concepts, such as drugs and genes, we filtered the entities that did not refer to diseases.
We refer to these as MedMentions-d.

\section{Hyperparameters}
At training, we randomly selected 16 concepts and extracted two sentences for each concept (i.e.,  32 sentences for each mini-batch).
Models were trained with the temperature parameter $\alpha$ set to 2, $\beta$ set to 50, the offset $\lambda$ set to 1, and the learning parameter set to 1e-5.

\section{Baseline Models}
\label{baseline}
The baseline entity representations used herein are as follows:

\begin{itemize}
    \item Tf-idf: Tf-idf of character level uni-grams and bi-grams. This achieves comparative performance as reported in previous works~\cite{Sung2020-ra}.
    \item PubMedBERT: The representations of \textit{[CLS]} from PubMedBERT without fine-tuning.
    \item \textsc{SapBert}: The representations of \textit{[CLS]} from PubMedBERT trained similar to that in \cite{Liu2020-mr}. Note that the size of the dictionary used in training and inference is equal to that used for our model, and is different from the original paper.
\end{itemize}

\section{Entity Tag}
\begin{table}[th]
\centering
\small
\begin{tabular}{lccc}
\toprule 
Methods & NCBID & BC5CDR & MedMentions \\
\midrule
BioCoM-all & 60.3 & 69.0 & 71.4\\
BioCoM-one & 59.3 & 66.9 & 70.7\\
\bottomrule
\end{tabular}
\caption{\label{result2} Entity tag ``all" implies that all of entities in the sentence are linked, and ``one" donates that only one entity is linked in the sentence.}
\end{table}

In practice, we need to identify disease mentions somehow (e.g. named entity recognition system).
As the database constructed with our method is only linked to the concepts in the dictionary, a the gap exists between the input sentence and database.
This may cause unexpected attention to the [\textit{ENT}] tags for the mentions that are not included in the dictionary.
One possible method to alleviate this problem is to link only one entity in a sentence.
If a sentence has multiple entities, we established linked sentences in which each entity is linked.
We trained and evaluated the model in this setting on MedMentions.

Table \ref{result2} shows the result.
The model using the sentences in which only one entity is linked performs poorly.
One possible reason for this is the small number of entities in the mini-batch.
Thus, number of negative samples is not enough for training.

\section{Effect of the Corpus Size}
\begin{figure}
    \centering
    \includegraphics[clip,width=7.0cm]{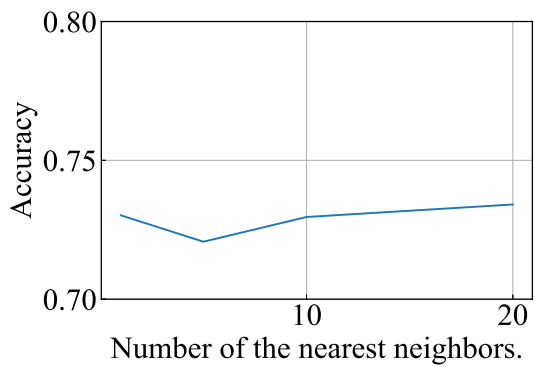}
    \caption{Number of sentences vs accuracy}
    \label{result3}
\end{figure}

When large numbers of documents are used for nearest neighbor search, using all sentences for normalization is memory-intensive.
We investigated how we can reduce memory usage for prediction.
One possible way is random sampling of the sentences in the database.
We evaluated the model with the sentences for each dictionary mention of 1, 5, 10, 20 on MedMentions-d.

Figure \ref{result3} shows the results.
Surprisingly, we found that the accuracy of the model with random sampling was higher than unity in all the sentences.
One possible reason for this is that dictionary names consisting of common words (e.g., cancer and tumor) are linked in many sentences.
These sentences provide harmful noise for our model.
Regarding the number of sentences, although the model with 20 sentences for each concept performed best, the model can normalize the mentions with a fewer number of sentences.
This implies that our model learns robust contextual entity representations.

\begin{table}[t!]
    \centering
    \footnotesize
    \begin{tabularx}{\linewidth}{lX}
        \toprule 

        Label: \green{D000077273} & Input: ... histopathology consisted of papillary thyroid carcinoma (\textbf{PTC}) (n = 91, 86.7\%).\\
        \hdashline
        \textsc{SapBert} & [\orange{C536943}] \textbf{TCC}\\
        \textsc{BioCoM} & [\green{D000077273}] ... for detecting cervical lymph node (LN) metastasis in \textbf{papillary thyroid carcinoma} (\textit{PTC}). \\
        
        \midrule
        Label: \green{D015430} & Input: ... generations on offspring metabolic traits , including \textbf{weight} and fat gain, ... \\
        \hdashline
        \textsc{SapBert} & [\green{D015430}] \textbf{gain, weight} \\
        Proposed & [\orange{D001835}] ... more elderly and had lower \textbf{weights , body} mass indices and arm and calf circumferences ... \\
        \bottomrule
\end{tabularx}
\caption{\label{other_example} More prediction samples from \textsc{SapBert} and \textsc{BioCoM}.
Each example shows the [predicted concept] and the nearest neighbors.
Entity mentions and the cue phrases are written in \textbf{bold} and \textit{italic} forms, respectively.}
\end{table}

\end{document}